\documentclass[letterpaper]{article} 
\usepackage{aaai24}  
\usepackage{times}  
\usepackage{helvet}  
\usepackage{courier}  
\usepackage[hyphens]{url}  
\usepackage{graphicx} 
\usepackage{natbib}  
\usepackage{caption} 
\usepackage{algorithm}
\usepackage{algorithmic}
\usepackage{amsmath}
\usepackage{amssymb}
\usepackage{url}
\usepackage{algorithm}
\usepackage{algorithmic}
\usepackage{multirow}
\usepackage{makecell}
\usepackage{booktabs}
\usepackage{subfigure}
\usepackage{subfloat}
\usepackage{lipsum}
\newtheorem{theorem}{Theorem}

\usepackage{soul}
\usepackage{xcolor}
\makeatletter
  \newcommand\figcaption{\def\@captype{figure}\caption}
  \newcommand\tabcaption{\def\@captype{table}\caption}
\makeatother
\usepackage{newfloat}
\usepackage{listings}
\DeclareCaptionStyle{ruled}{labelfont=normalfont,labelsep=colon,strut=off} 
\floatstyle{ruled}
\newfloat{listing}{tb}{lst}{}
\floatname{listing}{Listing}
\title{Combating Data Imbalances in Federated Semi-supervised Learning \\ with Dual Regulators}

\author{
    Sikai Bai\textsuperscript{\rm1}\equalcontrib,
    Shuaicheng Li\textsuperscript{\rm2}\equalcontrib\leader,
    Weiming Zhuang\textsuperscript{\rm3}\corresponding,\\
    Jie Zhang\textsuperscript{\rm4}\corresponding,
    Kunlin Yang\textsuperscript{\rm2},
    Jun Hou\textsuperscript{\rm2}\corresponding,
    Shuai Zhang\textsuperscript{\rm2},
    Shuai Yi\textsuperscript{\rm2},
    Junyu Gao\textsuperscript{\rm5}   
}
\affiliations {
    \textsuperscript{\rm 1}The Hong Kong University of Science and Technology \\
    \textsuperscript{\rm 2}SenseTime Research \\
    \textsuperscript{\rm 3}Sony AI\\
    \textsuperscript{\rm 4}The Hong Kong Polytechnic University \\
    \textsuperscript{\rm 5}Northwestern Polytechnical University \\
    \{whitesk1973, gjy3035\}@gmail.com,
    \{lishuaicheng, yangkunlin,  houjun, yishuai, zhangshuai\}@sensetime.com,
    weiming001@e.ntu.edu.sg,
    jie-comp.zhang@polyu.edu.hk
}

\usepackage{bibentry}

\begin{document}

\maketitle

\begin{abstract}
Federated learning has become a popular method to learn from decentralized heterogeneous data. Federated semi-supervised learning (FSSL) emerges to train models from a small fraction of labeled data due to label scarcity on decentralized clients. Existing FSSL methods assume independent and identically distributed (IID) labeled data across clients and consistent class distribution between labeled and unlabeled data within a client. This work studies a more practical and challenging scenario of FSSL, where data distribution is different not only across clients but also within a client between labeled and unlabeled data. To address this challenge, we propose a novel FSSL framework with dual regulators, FedDure. FedDure lifts the previous assumption with a coarse-grained regulator (C-reg) and a fine-grained regulator (F-reg): C-reg regularizes the updating of the local model by tracking the learning effect on labeled data distribution; F-reg learns an adaptive weighting scheme tailored for unlabeled instances in each client. We further formulate the client model training as bi-level optimization that adaptively optimizes the model in the client with two regulators. Theoretically, we show the convergence guarantee of the dual regulators. Empirically, we demonstrate that FedDure is superior to the existing methods across a  wide range of settings, notably by more than 11\% on CIFAR-10 and CINIC-10 datasets.
\end{abstract}

\section{Introduction}

Federated learning (FL) is an emerging privacy-preserving machine learning technique \cite{mcmahan2017communication}, where multiple clients collaboratively learn a model under the coordination of a central server without exchanging private data. 
It has empowered a wide range of applications, including healthcare \cite{kaissis2020secure,li2019brain-tumor1}, consumer products \cite{hard2018gboard,niu2020recommendation}, and etc.

\begin{figure}[t]
   \centering
   \includegraphics[width=0.85\linewidth]{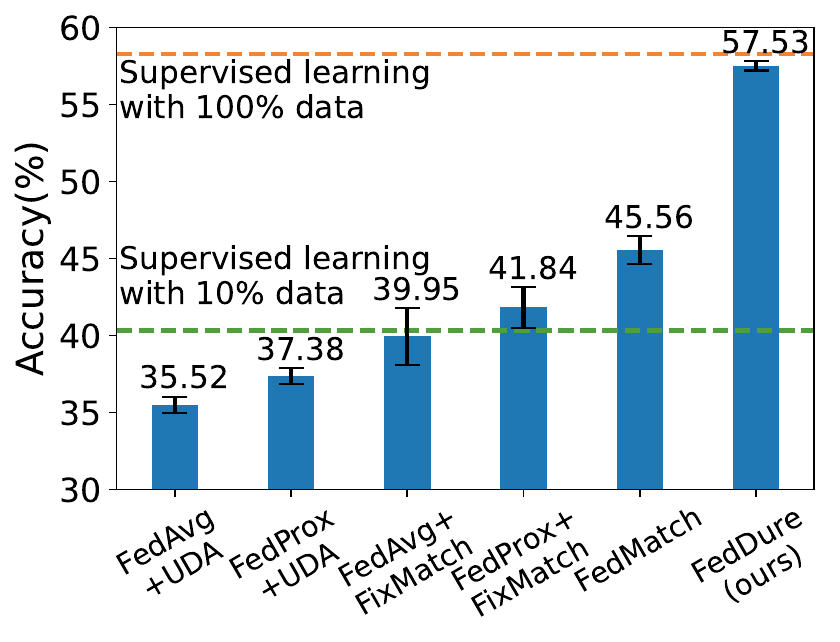}
   \caption[Caption for LOF]{Existing federated semi-supervised learning (FSSL) methods cannot address heterogeneity between labeled and unlabeled data within a client (internal imbalance) and heterogeneous data across clients (external imbalance); some of them are even worse than supervised FL using 10\% data (green line, which is FedAvg* in Table \ref{comparasion}). Our proposed FedDure significantly outperforms existing methods. These experiments are based on three runs on CIFAR-10 and we provide more description in Section Experiments.} 
   \label{fig1}
\end{figure}

The majority of existing FL works \cite{mcmahan2017communication, wang2020federated, li2021model} assume that the private data in clients are fully labeled, but the assumption is unrealistic in real-world federated applications as annotating data is time-consuming, laborious, and expensive. 
To remedy these issues, federated semi-supervised learning (FSSL) is proposed to improve model performance with limited labeled and abundant unlabeled data on each client \cite{jin2020survey}.
In particular, prior works~\cite{jeong2020federated,liu2021federated} have achieved competitive performance by exploring inter-client mutual knowledge.
However, they usually focus on mitigating heterogeneous data distribution across clients (\textbf{external imbalance}) while assuming that labeled and unlabeled training data are drawn from the same independent and identical distribution.
These assumptions enforce strict requirements of data annotation and would not be practical in many real-world applications.
A general case is that labeled and unlabeled data are drawn from different distributions (\textbf{internal imbalance}).
For example, photo gallery on mobile phones contains many more irrelevantly unlabeled images than the ones that are labeled manually for classification task \cite{2011yangIrrelevant}.

Existing FSSL methods perform even worse than training with only a small portion of labeled data, under this realistic and challenging FSSL scenario with external and internal imbalances, as shown in Figure \ref{fig1}. The main reasons of performance degradation are two-fold: 1) internal imbalance leads to intra-client skewed data distribution, resulting in heterogeneous local training; 2) external imbalance leads to inter-client skewed data distribution, resulting in client drift \cite{charles2021convergence, karimireddy2020scaffold}. The co-occurrence of internal and external data imbalances amplifies the impact of client drifts and local inconsistency, leading to performance degradation.

To address the above issues, we propose a new federated semi-supervised learning
framework termed \textbf{FedDure}.
FedDure explores two adaptive regulators, a coarse-grained regulator (\textbf{C-reg}) and a fine-grained regulator (\textbf{F-reg}), to flexibly update the local model according to the learning process and outcome of the client's data distributions.
Firstly, C-reg regularizes the updating of the local model by tracking the learning effect on labeled data. 
By utilizing the real-time feedback from C-reg, FedDure rectifies inaccurate model predictions and mitigates the adverse impact of internal imbalance. 
Secondly, F-reg learns an adaptive weighting scheme tailored for each client; it automatically equips a \textit{soft} weight for each unlabeled instance to measure its contribution.
This scheme automatically adjusts the instance-level weights to strengthen (or weaken) its confidence according to the feedback of F-reg on the labeled data to further address the internal imbalance. 
Besides, FedDure mitigates the client drifts caused by external imbalance by leveraging the global server model to provide guidance knowledge for C-reg. 
During the training process, FedDure utilizes the bi-level optimization strategy to alternately update the local model and dual regulators in local training. Figure \ref{fig1} shows that FedDure significantly outperforms existing methods and its performance is even close to fully supervised learning (orange line) under internal and external imbalance. To the end, the main contributions are three-fold:

\begin{itemize}
\item We are the first work that investigates a more practical and challenging scenario of FSSL, where data distribution differs not only across clients (external imbalance) but between labeled and unlabeled data within a client (internal imbalance).
    \item We propose FedDure, a new FSSL framework that designs dual regulators to adaptively update the local model according to the unique learning processes and outcomes of each client.
    \item We theoretically analyze the convergence of dual regulators and empirically demonstrate that FedDure is superior to the state-of-the-art FSSL approaches across
    multiple benchmarks and data settings, improving accuracy by 12.17\% on CIFAR10 and by 11.16\% on CINIC-10 under internal and external imbalances.
\end{itemize}

\begin{figure*}[t]
	\begin{center}
		\centering
	\includegraphics[width=0.90\linewidth]{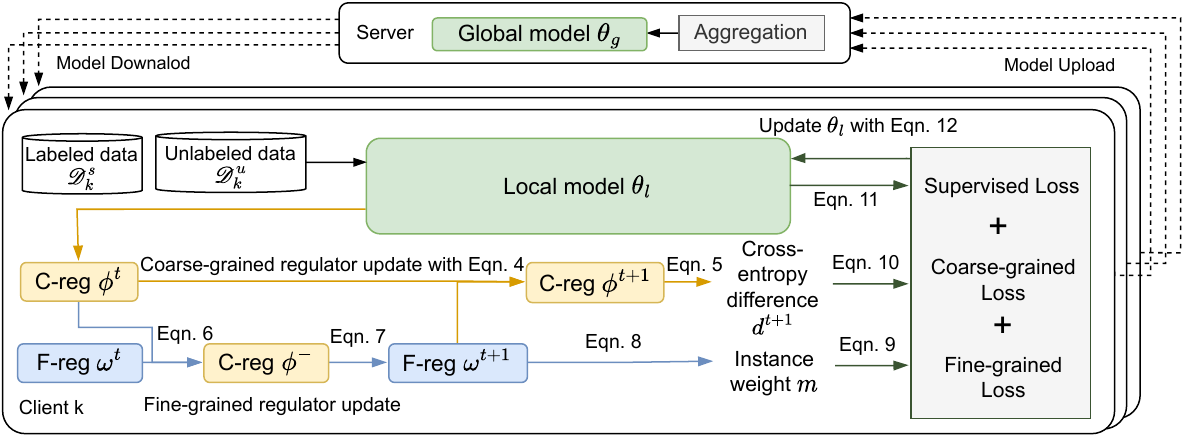}
		\hfill
	\end{center}
	\caption{Illustration of \underline{Fed}erated Semi-Supervised Learning Framework with \underline{Du}al \underline{Re}gulator (FedDure). FedDure contains a coarse-grained regulator (C-reg) and a fine-grained regulator (F-reg) to adaptively guide local updates in each client: C-reg dynamically regulates the importance of local training on the unlabeled data by reflecting the overall learning effect on labeled data; F-reg regulates the performance contribution of each unlabeled sample.}\label{fig:main}
\end{figure*}

\section{Related Work}

\textbf{Federated Learning (FL)} is an emerging distributed training technique that trains models on decentralized clients and aggregates model updates in a central server  \cite{yang2019federated}. 
FedAvg \cite{mcmahan2017communication} is a pioneering work that aggregates local models updated by weighted averaging. Statistical heterogeneity is an important challenge of FL in real-world scenarios, where the data distribution is inconsistent among clients \cite{li2020federated}, which can result in drift apart between global and local model, i.e., client-drift \cite{charles2021convergence}. 
A plethora of works have been proposed to address this challenge with approaches like extra data sharing, regularization, new aggregation mechanisms, and personalization \cite{zhao2018federated, li2022probing, li2021fedbn, xu2021fedcm}. These approaches commonly consider only supervised learning settings and may not be simply applied to scenarios where only a small portion of data is labeled. 
Moreover, some work studies un/self-supervised learning settings
\cite{zhuang2022fessl, wang2021unsupervised, li2021groupformer} to learn generic representations with purely unlabeled data on clients, and these methods require IID labeled data for fine-tuning the representations for downstream tasks \cite{li2022pyramid, bai2021multi}.
Our work primarily focuses on federated semi-supervised learning, where a small fraction of data has labels in each client.

\textbf{Semi-Supervised Learning} aims to utilize unlabeled data for performance improvements and is usually divided into two popular branches pseudo labeling and consistency regularization.
Pseudo-labeling methods \cite{lee2013pseudo, zou2021pseudoseg,pham2021meta,DST} usually generate artificial labels of unlabeled data from the model trained by labeled data and apply the filtered high-confidence labels as supervised signals for unlabeled data training.
MPL \cite{pham2021meta} extends the knowledge distillation to SSL by optimizing the teacher model with feedback from the student model.
Consistency regularization approaches \cite{lee2022contrastive, tarvainen2017mean} regularize the outputs of different perturbed versions of the same input to be consistent.
Many works \cite{sohn2020fixmatch,zhang2021flexmatch,lee2022contrastive} apply data augmentation as a perturbed strategy for pursuing outcome consistency. 

\textbf{Federated Semi-Supervised Learning (FSSL)} considers learning models from decentralized clients where a small amount of labeled data resides on either clients or the server \cite{jin2020towards}. FSSL scenarios can be classified into three categories: (1) Labels-at-Server assumes that clients have purely unlabeled data and the server contains some labeled data \cite{lin2021semifed, he2021ssfl, zhang2021improving, diao2021semifl}; (2) Labels-at-Clients considers each client has mostly unlabeled data and a small amount of labeled data \cite{jeong2020federated}; (3) Labels-at-Partial-Clients assumes that the majority of clients contain fully unlabeled data while numerous clients have fully labeled data \cite{lin2021semifed, liang2022rscfed}. Labels-at-Clients has been largely overlooked; 
prior work \cite{jeong2020federated} proposes inter-client consistency loss, but it shares extra information among clients and bypasses the internal class imbalance issue. 
This work introduces dual regulators to address the issue, without extra information shared among clients.


\section{Method}
This section first defines the problem and introduces a novel framework with dual regulators (FedDure). Using dual regulators, we then build a bi-level optimization strategy for federated semi-supervised learning.

\subsection{Problem Definition}
We focus on Federated Semi-Supervised Learning (FSSL) with external and internal imbalance problems. Specifically,
we assume that there are $K$ clients, denoted as $\{ \mathcal{C}_1, ..., \mathcal{C}_K\}$. Federated learning aims to train a generalized global model $f_g$ with parameter $\theta_g$. It coordinates decentralized clients to train their local models $\mathcal{F}_l = \{ f_{l,1}, ...,f_{l,K} \}$ with parameters $\{ \theta_{l,1}, ...,\theta_{l,K} \}$, where each client is only allowed to access its own local private dataset. In the standard semi-supervised setting, the dataset contains a labeled set $\mathcal{D}^s = \{ \textbf{x}_i, \textbf{y}_i\}_{i=1}^{N^s}$ and an unlabeled set $\mathcal{D}^u = \{ \textbf{u}_i\}_{i=1}^{N^u}$, where $N^s \ll N^u$. Under FSSL, the private dataset $\mathcal{D}_k$ of each client $C_k$ contains $N_k^s$ labeled instances $\mathcal{D}^s_k = \{ \textbf{x}_{i,k}, \textbf{y}_{i,k}\}_{i=1}^{N^s_k}$ and $N_k^u$ unlabeled instances $\mathcal{D}^u_k = \{ \textbf{u}_{i,k}\}_{i=1}^{N^u_k}$. The internal imbalance means that the distribution of $\mathcal{D}^s_k$ and $\mathcal{D}^u_k$ are different; the external imbalance refers to different distributions between $D_k$ in different clients $k$. We provide a detailed description in Subsection Data Heterogeneity. 

In this work, we primarily focus on image datasets. For an unlabeled image $\textbf{u}_k$ in client $C_k$, we compute the corresponding pseudo label $\hat{\textbf{y}}_k$ with the following equation:
\begin{equation}
	\hat{\textbf{y}}_k = \mathrm{argmax} (f_{l,k}(\mathcal{T}_w( \textbf{u}_k); \pmb{\theta}_{l,k})),
	\label{eq1}
\end{equation}
where $\mathcal{T}_w( \textbf{u}_k)$ is the weakly-augmented version of $\textbf{u}_k$ and the pseudo labeling dataset in the client $C_k$ is denoted as  $\mathcal{D}^u_k = \{ \textbf{u}_{i,k}, \hat{\textbf{y}}_{i,k}\}_{i=1}^{N_{k}^u}$. {We omit the client index $k$ in the parameters later for simplicity of notation.}

\subsection{Dual Regulators}
In this section, we present \underline{fed}erated semi-supervised learning with \underline{du}al \underline{re}gulator, termed FedDure. 
It dynamically adjusts gradient updates in each client according to the class distribution characteristics with two regulators, a coarse-grained regulator (C-reg) and a fine-grained regulator (F-reg). 
Figure \ref{fig:main} depicts the optimization process with these two regulators. We introduce the regulators and present the optimization process in the following subsections. 

\textbf{Coarse-grained Regulator (C-reg).} \; 
Existing FSSL methods decompose the optimization on the labeled and unlabeled data, leading to heterogeneous local training. C-reg remedies the challenge with a collaborative training manner. Intuitively,
the parameters of the local model can be rectified according to the feedback from C-reg, which dynamically regulates the importance of local training on all unlabeled data by quantifying the overall learning effect using labeled data.
It contributes to counteracting the adverse impact introduced by internal imbalance and preventing corrupted pseudo-labels \cite{DST}. Meanwhile, C-reg acquires global knowledge by initializing with the received server model parameters at the beginning of each round of local training, which can provide global guidance to the local model to mitigate external imbalance (client-drift).

We define C-reg as $f_d$ with parameters $\pmb{\phi}$. 
At training iteration $t$, C-reg searches its optimal parameter $\pmb{\phi}^*$  by minimizing the cross-entropy loss on unlabeled data with pseudo labels.
Actually, the optimal parameter $\pmb{\phi}^*$ is related to the local model's parameter $\pmb{\theta}_l$ via the generated pseudo label, and we denote the relationship as $\pmb{\phi}^*(\pmb{\theta}_l)$. 
Since it requires heavy computational costs to explore the optimal parameter $\pmb{\phi}^*$, we approximate $\pmb{\phi}^*$ by performing \textit{one gradient step} $\pmb{\phi}^{t+1}$  at training iteration $t$ (i.e., $\pmb{\phi}^{t}$).

Practically, we introduce the updated fine-grained regulator (F-reg) to measure the scalar weight for each unlabeled instance for updating C-reg.
The formulation to optimize C-reg is as follows:
\begin{equation}
\begin{split}
\label{eq_phi}
	 \pmb{\phi}^{t+1} \! =\! \pmb{\phi}^{t} \! -\! \eta_s   {
	 \!  \nabla_{\pmb{\phi}^t} \mathbb{E}_{\pmb{u}} \mathcal{H} (\pmb{w}^{t+1};\pmb{\phi}^{t}) \mathcal{L}_{ce} \left(\hat{\textbf{y}}, f_d\left(\mathcal{T}_s(\textbf{u}); \pmb{\phi}^{t}\right)\right)},
\end{split}
\end{equation}
where $\mathcal{H} (\pmb{w}^{t+1};\pmb{\phi}^{t}) = f_w\left(f_d\left(\mathcal{T}_s(\textbf{u}); \pmb{\phi}^{t}\right);\pmb{w}^{t+1}\right)$, $f_w$ is the fine-grained regulator (F-reg), and $\pmb{w}^{t+1}$ is the parameters of F-reg updated by Eqn. \ref{eq6}, which is detailed in the following subsection. $\mathcal{T}_s(\textbf{u})$ is the strongly-augmented unlabeled image $\textbf{u}$ and $f_d(\mathcal{T}_s(\textbf{u}); \pmb{\phi}^{t})$ is the output vector of $f_d$ to evaluate the quality of pseudo labels from the local model.

Next, we quantify the learning effect of the local model with the C-reg using labeled samples by computing the cross-entropy difference $d^{t+1}$ of C-reg between training iterations $t$ and $t+1$:
\begin{equation}
	d^{t+1} \! = \! \mathbb{E}_{\textbf{x},\textbf{y}} \left[  {\mathcal{L}_{ce} (
    \textbf{y}, f_d(\textbf{x}; \pmb{\phi}^{t})}) \!- \!{\mathcal{L}_{ce} (\textbf{y}, f_d(\textbf{x}; \pmb{\phi}^{t+1} )} )\right]. 
	\label{eq_entropy_difference}
\end{equation}
The quantized learning effect is further used as the reward information to optimize the local model by regulating the importance of local training on unlabeled data. In particular, the cross-entropy differences $d^{t+1}$ signify the generalization gap for the C-reg updated by the pseudo labels from the local model. 

\textbf{Fine-grained Regulator (F-reg).} \; 
Previous SSL methods usually utilize a \textit{fixed threshold} to filter noisy pseudo labels \cite{sohn2020fixmatch}, but they are substantially hindered by corrupted labels or class imbalance on unlabeled data.
Internal and external imbalances in FSSL could amplify these problems, leading to 
performance degradation.
To tackle the challenge, F-reg regulates the importance of each unlabeled instance in local training for mitigating the learning bias caused by internal imbalance.
It learns an adaptive weighting scheme tailored for each client according to unlabeled data distribution. A unique weight is generated for each unlabeled image to measure the contribution of the image to overall performance. 
We construct F-reg $f_w$ parameterized by $\pmb{w}$\footnote{F-reg is a MLP architecture with one fully connected layer with 128 filters and a Sigmoid function.}. Before updating F-reg, we perform \textit{one gradient step} update of C-reg $\pmb{\phi}$ to associate F-reg and C-reg:
\begin{equation}
 \pmb{\phi}^{-}\! = \! \pmb{\phi}^{t} \! -\! \eta_s  {  
	  \nabla_{\pmb{\phi}^t} \mathbb{E}_{\pmb{u}} \mathcal{H} (\pmb{w}^{t};\pmb{\phi}^{t}) \mathcal{L}_{ce}\left(\hat{\textbf{y}}, f_d\left(\mathcal{T}_s(\textbf{u}); \pmb{\phi}^{t}\right)\right)},
\label{eq_phi_new}
\end{equation}
where \textit{one gradient step} of C-reg $\pmb{\phi}^{-}$ depends on the F-reg $\pmb{w}^{t}$ and regards the others as fixed parameters. Next, we optimize F-reg in local training iteration $t$, where the optimal parameter $\pmb{w}^* $ is approximated by one gradient step of F-reg (i.e., $\pmb{w}^{t+1}$). The optimization of F-reg is formulated as:
\begin{equation}
\begin{split}  
 \pmb{w}^{t+1} = \pmb{w}^{t} - \eta_w  \nabla_{\pmb{w}^{t}} \mathbb{E}_{\textbf{x},\textbf{y}} \mathcal{L}_{ce} \left(\textbf{y},{f}_d(\textbf{x}; \pmb{\phi^{-}}(\pmb{w}^{t})\right),
\label{eq6}
\end{split}
\end{equation}
where ${f}_d(\textbf{x}; \pmb{\phi}^{-}(\pmb{w}^{t}))$ is the output of $f_d$ on labeled data. We then introduce a re-weighting scheme that calculates a unique weight $\pmb{m}_i$ for $i$-th unlabeled sample: 
\begin{equation}
	\pmb{m}_i = f_w({f}_l(\mathcal{T}_s(\textbf{u}_i), \pmb{\theta}_l^{t}), \pmb{w}^{t+1}).
\label{eq_m}
\end{equation}
Note that $\pmb{m}_i$ is a scalar to re-weight the importance of the corresponding unlabeled image.

\subsection{Bi-level Optimization}
\label{sec:bi-level-optim}

In this section, we present optimization processes for the dual regulators and local model $\theta$. 
We alternatively train two regulators, which approximate a gradient-based bi-level optimization procedure \cite{finn2017model_MAML,liu2018darts}. Then, we update the local model with fixed C-reg and F-reg.

\textbf{Update F-reg.} Firstly, we obtain one gradient step update of C-reg $ \pmb{\phi}^{-}$ using Eqn. \ref{eq_phi_new}. After that, the supervised loss $\mathcal{L}_{ce} \left( \textbf{y},{f}_d(\textbf{x}; \pmb{\phi}^{-}(\pmb{w}^{t})\right)$ guides the update of the F-reg with Eqn. \ref{eq6}.
Since $\pmb{w}^{t}$ is explicitly beyond the supervised loss, the updating of F-reg can be achieved by standard backpropagation using the chain rule.

\textbf{Update C-reg.}
After updating the parameters of F-reg, we update C-reg by Eqn. \ref{eq_phi}, regarding local model $\pmb{\theta_l}^{t}$ as fixed parameters.

\textbf{Update Local Model with F-reg.}
We use the updated F-reg $\pmb{w}^{t+1}$ to calculate a unique weight $\pmb{m}_i$ for $i$-th unlabeled sample with Eqn. \ref{eq_m}.
The gradient optimization is formulated as:
\begin{equation}
	\pmb{g}^{t}_u =     \mathbb{E}_{\textbf{u}} \left[	 \nabla_{\pmb{\theta}^{t}_l}  \mathcal{L}_{ce}\left(\hat{\textbf{y}}, {f}_l\left(\mathcal{T}_s(\textbf{u}); \pmb{\theta}_l^{t}\right)\right) \cdot \pmb{m} \right].
	\label{eq_u_loss}
\end{equation}

\textbf{Update Local Model with C-reg.} We then use C-reg to calculate entropy difference $d^{t+1}$ in Eqn. \ref{eq_entropy_difference}. The entropy difference $d^{t+1}$ is adopted as a reward coefficient to adjust the gradient update of the local model on unlabeled data. The formulation is as follows:
\begin{equation}
	\pmb{g}^{t}_d = d^{t+1} \cdot \nabla_{\pmb{\theta}^{t}_l} \mathbb{E}_{\textbf{u}} \mathcal{L}_{ce}\left(\hat{\textbf{y}}, f_l\left(\mathcal{T}_s(\textbf{u}); \pmb{\theta}_l^{t}\right)\right),
	\label{eq_d_loss}
\end{equation}
where this learning process is derived from a meta-learning strategy, provided in supplementary materials about the proof for analysis. 

\textbf{Update Local Model with Supervised Loss.} Besides, we compute the gradient local model on labeled data as:
\begin{equation}
	\pmb{g}^{t}_s =   \nabla_{\pmb{\theta}^{t}_l}  \mathbb{E}_{\textbf{x},\textbf{y}} \mathcal{L}_{ce}\left(\textbf{y}, f_l\left(\textbf{x}; \pmb{\theta}_l^{t}\right)\right).
	\label{eq_supervised}
\end{equation}

On this basis, we update the local model’s parameter with the above gradient computation in Eqn. \ref{eq_u_loss}, \ref{eq_d_loss} and \ref{eq_supervised}, which is defined as:
\begin{equation}
	\pmb{\theta}^{t+1}_l =\pmb{\theta}^{t}_l - \eta \left(  \pmb{g}^{t}_s  +   \pmb{g}^{t}_u +  \pmb{g}^{t}_d\right),
	\label{eq_total_gradient}
\end{equation}
where $\eta$ denotes the learning rate of the local model.
Finally, after $T$ local epochs, the local model is returned to the central server.  The server updates global model $\pmb{\theta_g}^{r+1}$ by weighted averaging the parameters from these received local models in the current round, and $r+1$-th round is conducted by sending $\pmb{\theta_g}^{r+1}$ to the randomly selected clients as initialization. We present the pipeline of the overall optimization process in the supplementary material about the proof for analysis.

\subsection{Convergence of Optimization Process}
In this section, we further analyze the convergence of our optimizations.
When updating F-reg in Eqn. \ref{eq6}, $\pmb{w}^t$ is explicitly beyond the supervised loss, the optimization of F-reg can be easily implemented by automatic backpropagation using the chain rule.
We only discuss the convergence of the bi-level optimizations using the meta-learning process.

\noindent \textbf{Update Local Model with C-reg.}
The local model tries to update its parameters on the feedback from the updated coarse-grained regulator (C-reg), which adjusts the learning effect via the meta-learning process. 
The cross-entropy loss on labeled data $\mathcal{L}_{ce}(\textbf{y}, f_d(\textbf{x}; \pmb{\phi}^{t+1}(\pmb{\theta}_l^{t}))$ is applied to characterize the quality of learning effect from the local model. The CE loss function is related to $\pmb{\theta}_l^{t}$.

\begin{theorem} \label{th_theorem_1}
{Suppose that supervised loss function} $\mathcal{L}_{ce}(\textbf{y}, f_d(\textbf{x}; \pmb{\phi}^{t+1}(\pmb{\theta}_l^{t}))$ is $L$-Lipschitz and has $\rho$-bounded gradients. 
The $\mathcal{L}_{ce}\left(\hat{\textbf{y}}, f_d\left(\mathcal{T}_s(\textbf{u}); \pmb{\phi}^{t}\right)\right)$ has $\rho$-bounded gradients and twice differential with Hessian bounded by $\mathcal{B}$.
Let the learning rate $\eta_s=\min \{ 1, \frac{e}{T} \}$ for constant $e > 0$, and $\eta= \min \{ \frac{1}{L}, \frac{c}{\sqrt{T}} \}$ for some $c > 0$, such that $\frac{\sqrt{T}}{c} \geq L$. 
Thus, the optimization of the local model using coarse-grained regulator can achieve:
\begin{equation}
    \begin{aligned}
		\min_{0\leq t \leq T} \mathbb{E}[ \|\nabla_{\theta_l} \mathcal{L}_{ce}(\textbf{y}, f_d(\textbf{x}; \pmb{\phi}^{t+1}(\pmb{\theta}_l^{t}))\|_2^2] 
		\leq
		\mathcal{O}(\frac{c}{\sqrt{T}}).
    \end{aligned}
\end{equation}

\end{theorem}

\noindent\textbf{Update C-reg.}
We introduce updated F-reg to measure the contributions of each instance for updating C-reg in Eqn. \ref{eq_phi}, where $\mathcal{H} (\pmb{w}^{t+1};\pmb{\phi}^{t})$ is related to $\pmb{\phi}^t$. 
The updated F-reg adjusts the learning contributions on each unlabeled instance for regulating the optimization of C-reg. 
We conclude that our C-reg can always achieve convergence when introducing the feedback from F-reg. 

\begin{theorem} \label{th_theorem_2}
Suppose supervised and unsupervised loss functions are Lipschitz-smooth with constant L and have $\rho$-bounded gradient.
The $\mathcal{H}(\cdot)$ is differential with a $\epsilon$-bounded gradient and twice differential with its Hessian bounded by $\mathcal{B}$.
Let learning rate $\eta_s$ satisfies $\eta_s =\min \{ 1, \frac{k}{T} \}$ for constant $k > 0$, such that $\frac{k}{T} <1$.
$\eta_w = \min \{ \frac{1}{L}, \frac{c}{\sqrt{T}} \}$ for constant $c>0$ such that $\frac{\sqrt{T}}{c}\geq L$. 
The optimization of the coarse-grained regulator can achieve: 
\begin{equation}
\begin{split}
    \begin{aligned}
    \lim_{t \rightarrow \infty} \mathbb{E}[ \|{  
	  \nabla_{{\phi}}  \mathcal{H} (\pmb{w}^{t+1};\pmb{\phi}^{t}) \mathcal{L}_{ce}\left(\hat{\textbf{y}}, f_d\left(\mathcal{T}_s(\textbf{u}); \pmb{\phi}^{t}\right)\right) }\|_2^2]=0.
    \end{aligned}
\end{split}
\end{equation}
\end{theorem}

\begingroup
\setlength{\tabcolsep}{0.15em}
\begin{table*}[h]
	\begin{center}
		
		\resizebox{1.0\hsize}{!}{
			\begin{tabular}{c|c|c| c| c|c| c| c|c| c}
				\toprule[1.3pt]
				\multirow{2}{*}{Methods}   & \multicolumn{3}{c|}{CIFAR10} & \multicolumn{3}{c|}{Fashion-MNIST} & \multicolumn{3}{c}{CINIC-10}\\
				\cline{2-10} 
				  & {(IID, IID)} & {(IID, DIR)} & {(DIR, DIR)} & {(IID, IID)} & {(IID, DIR)} & {(DIR, DIR)} & {(IID, IID)} & {(IID, DIR)} & {(DIR, DIR)} \\
				\hline \hline
                    FedAvg* &45.68 &43.83 &40.34  &85.56 &84.84 &{82.24} &{40.73} &{39.00} &{28.09} \\
				FedAvg-SL  &{75.47} &{66.70} &{58.38} &{89.87} &{88.60} &{86.95} &{67.97} &{57.72} &{46.21}  \\
				FedProx-SL   &{74.67} &{66.78} &{59.55} &{89.53} &{88.35} &{87.32 } &{68.13} &{58.67} &{52.09} \\ \hline
                 FedAvg+UDA  &{47.47} &{43.89} &{35.52}  &{86.20} &{85.35} &{81.07} &{42.25 } &{39.93} &{29.27}  \\
                 FedProx+UDA  &{46.49} &{42.82} &{37.38} &{84.78} &{84.50} &{82.94 } &{41.81} &{39.40} &{33.26}  \\
				 FedAvg+Fixmatch  &{46.71} &{45.58} &{39.95} &{86.46} &{85.42} &{81.07} &{40.40} &{39.66} &{31.99}  \\
                  FedProx+Fixmatch  &{47.60} &{43,39} &{41.85} &{86.31} &{85.18} &{83.68} &{41.46} &{40.02} &{32.21}  \\
				FedMatch &{51.52} &{51.59} &{45.56} &{85.71} &{85.55} &{85.13} &{43.73} &{41.82} &{35.27}  \\
				\textbf{FedDure (Ours)} &\textbf{{67.69}} &\textbf{{66.85}} &\textbf{{57.73}} &\textbf{{88.69}} &\textbf{{88.21}} &\textbf{{86.96}} &\textbf{{56.36}} &\textbf{{55.10}} &\textbf{{46.43}}\\ 		
				\bottomrule[1.3pt]
			\end{tabular}
		}
  \tabcaption{Performance comparison of our proposed FedDure with state-of-the-art methods on three different data heterogeneity settings. FedDure outperforms all other methods in all settings.}\label{comparasion}
	\end{center}
 
\end{table*}
\endgroup

\begin{figure*}[t]

	\centering
    \subfigure[\small FedMatch: Labeled]{	
		\begin{minipage}[t]{0.22\linewidth}
			\centering
			\label{fig_dist_a}
			\includegraphics[width=1.0\columnwidth]{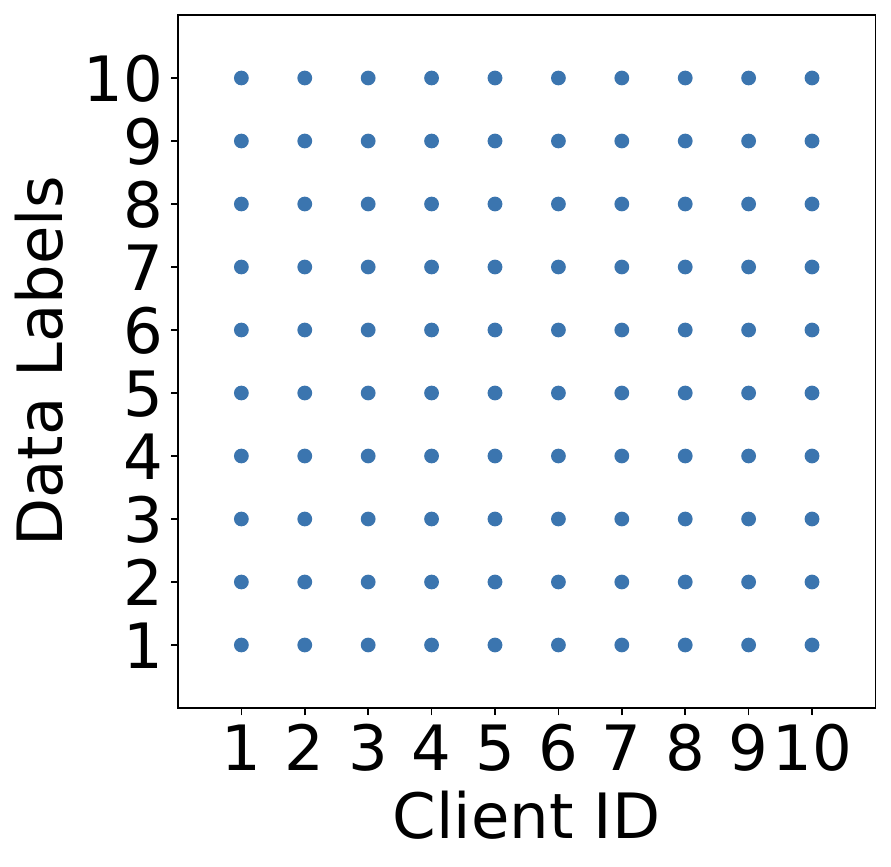}
		\end{minipage}%
	}%
	\hfill
    \subfigure[\small FedMatch: Unlabeled]{	
		\begin{minipage}[t]{0.22\linewidth}
			\centering
			\label{fig_dist_b}
			\includegraphics[width=1.0\columnwidth]{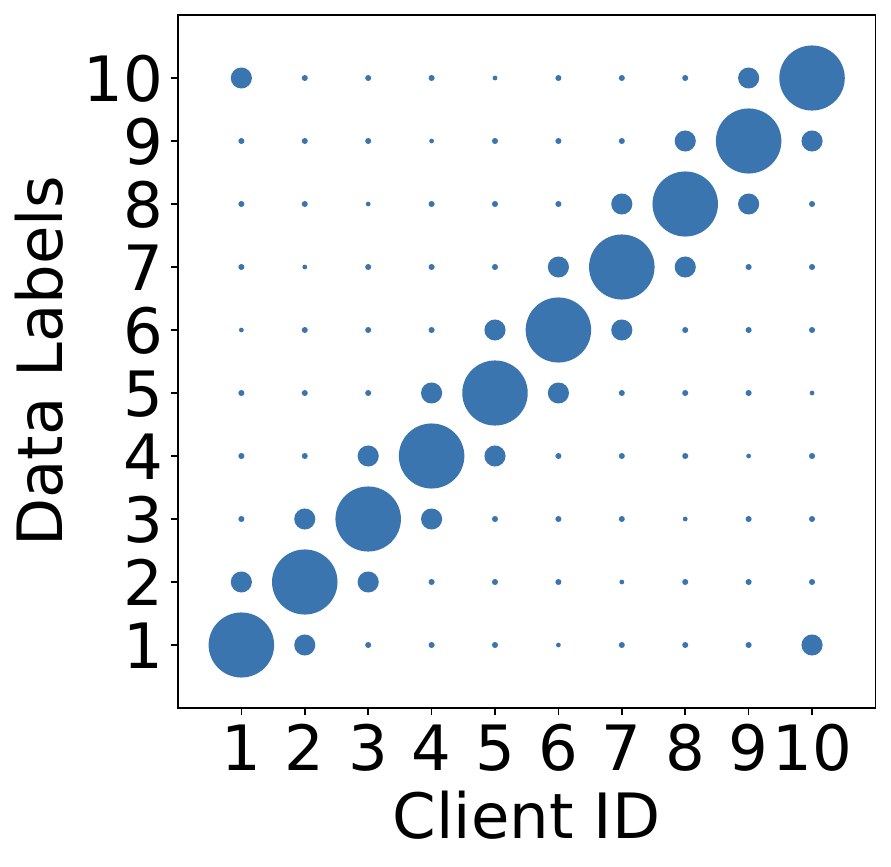}
		\end{minipage}%
	}
	\hfill%
    \subfigure[FedDure: Labeled]{
		\begin{minipage}[t]{0.22\linewidth}
			\centering
			\label{fig_dist_c}
			\includegraphics[width=1.0\columnwidth]{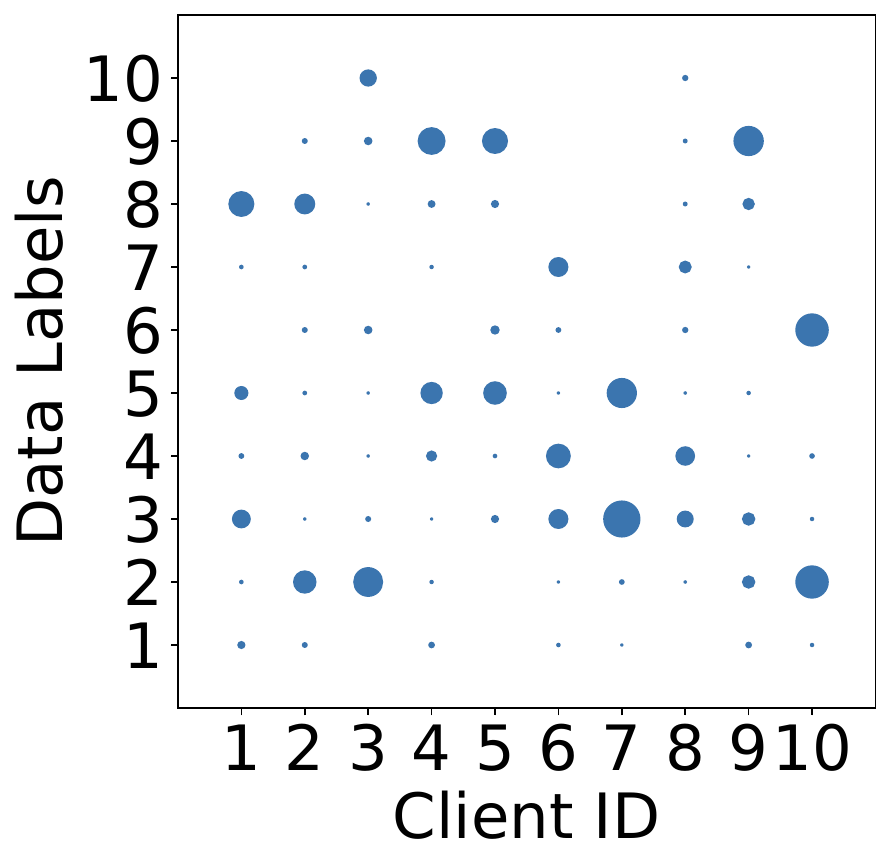}
		\end{minipage}%
	}%
	\hfill
    \subfigure[FedDure: Unlabeled]{
		\begin{minipage}[t]{0.22\linewidth}
			\centering
			\label{fig_dist_d}
			\includegraphics[width=1.0\columnwidth]{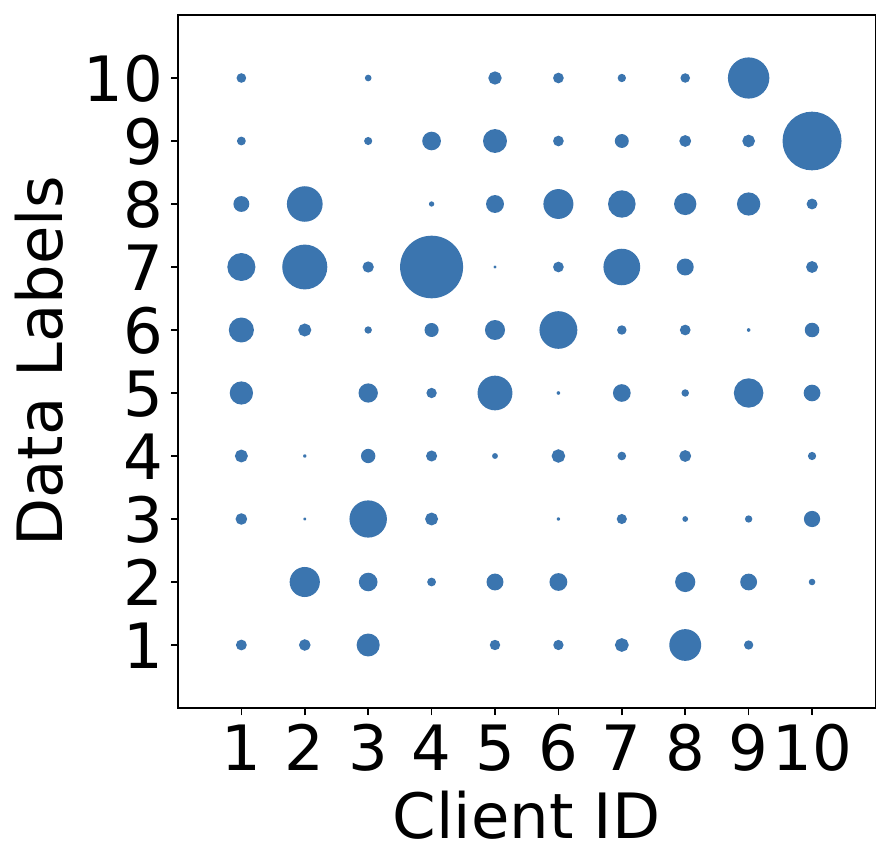}
		\end{minipage}%
	}%
	\centering
	\caption{Comparison of data distribution between FedMatch \cite{jeong2020federated} and our (DIR, DIR) setting: (a) and (b) are labeled and unlabeled data distribution used in FedMatch, respectively; our data distribution in (c) and (d) present external imbalance across clients and internal imbalance between labeled and unlabeled data inside a client.}
	\label{fig_dist}
\end{figure*}

\section{Experiments}
\label{experiments}

In this section, we demonstrate the effectiveness and robustness of our method through comprehensive experiments in three benchmark datasets under multiple data settings.

\subsection{Experimental Setup}
\label{sec:experimental-setup}

\textbf{Datasets.} \ We conduct comprehensive experiments on three datasets, including CIFAR-10 \cite{krizhevsky2009learning}, Fashion-MNIST \cite{xiao2017fashion} and CINIC-10 \cite{darlow2018cinic}. 
All datasets are split according to official guidelines; we provide more dataset descriptions and split strategies in the supplementary material. 

\textbf{Data Heterogeneity.} \ We construct three data heterogeneity settings with different data distributions. We denote each setting as \textbf{($\mathcal{A}$, $\mathcal{B}$)}, where $\mathcal{A}$ and $\mathcal{B}$ are data distribution of labeled and unlabeled data, respectively. The settings are as follows:
(1) \textit{\textbf{(IID, IID)}} means both labeled and unlabeled data are IID. By default, we use 5 instances per class to build the labeled dataset for each client. The remaining instances of each class are divided into $K$ clients evenly to build an unlabeled dataset.
(2) \textit{\textbf{(IID, DIR)}} means labeled data is the same as (IID, IID), but the unlabeled data is constructed with Dirichlet distribution to simulate data heterogeneity, where each client could only contain a subset of classes.
(3) \textit{\textbf{(DIR, DIR)}} constructs both labeled and unlabeled data with Dirichlet distribution. It simulates external and internal class imbalance, where the class distributions across clients and within a client are different. We allocate 500 labeled data per class to 100 clients using the Dirichlet process. The rest instances are divided into each client with another Dirichlet distribution. Figure \ref{fig_dist} compares the data distribution of FedMatch (Batch NonIID) \cite{jeong2020federated} and ours. 
Our (DIR, DIR) setting presents class imbalance across clients (external imbalance) and between labeled and unlabeled data within a client (internal imbalance).

\textbf{Implementation Details.} \; We use the Adam optimizer with momentum $=0.9$, batch size $=10$ and learning rates $=0.0005$ for $\eta_s$, $\eta$ and $\eta_w$. If there is no specified description, our default settings also include local iterations $T=1$, the selected clients in each round $S=5$, and the number of clients $K=100$. For the DIR data configuration, we use a Dirichlet distribution $Dir(\gamma)$ to generate the DIR data for all clients, where $\gamma=0.5$ for all three datasets. We adopt the ResNet-9 network as the default backbone architecture for local models and the coarse-grained regulator, while an MLP is utilized for the fine-grained regulator. 

\textbf{Baselines.} We compare the following methods in experiments. \textit{\textbf{FedAvg*}} denotes FedAvg \cite{mcmahan2017communication} only trained on labeled samples in FSSL (about 10\% data). \textit{\textbf{FedAvg-SL}} and \textit{\textbf{FedProx-SL}} are fully supervised training using FedAvg \cite{mcmahan2017communication} and FedProx \cite{li2020federatedop}, respectively.
\textit{\textbf{FedAvg+UDA}}, \textit{\textbf{FedProx+UDA}}, \textit{\textbf{FedAvg+Fixmatch}}, and \textit{\textbf{FedProx+Fixmatch}}: a naive combination between semi-supervised methods 
(UDA \cite{xie2020unsupervised} and Fixmatch \cite{sohn2020fixmatch}) and FL algorithms. 
They use labeled and unlabeled data, but  need to specify a predefined threshold on pseudo labels. 
\textit{\textbf{FedMatch}} \cite{jeong2020federated} adopts inter-consistency loss, 
and disjoint loss for model training
which is the state-of-the-art FSSL method. Note that we use the same hyper-parameters for FedDure and other methods in all experiments. 

\begin{figure}[t]
	\centering
    \subfigure[ Setting: (IID, DIR)]{	
		\begin{minipage}[t]{0.485\linewidth}
			\centering
			\label{fig_dir_IIDDIR}
			\includegraphics[width=1.0\columnwidth]{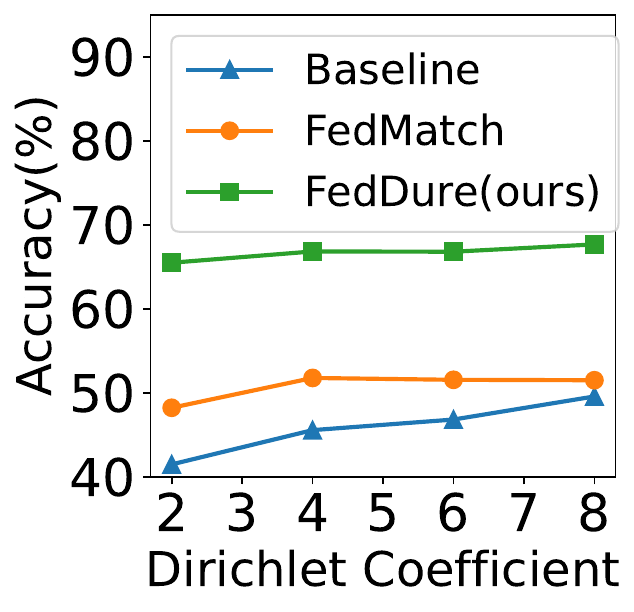}
		\end{minipage}%
	}%
	\hfill
    \subfigure[Setting: (DIR, DIR)]{	
		\begin{minipage}[t]{0.49\linewidth}
			\centering
			\label{fig_dir_DIRDIR}
			\includegraphics[width=1.0\columnwidth]{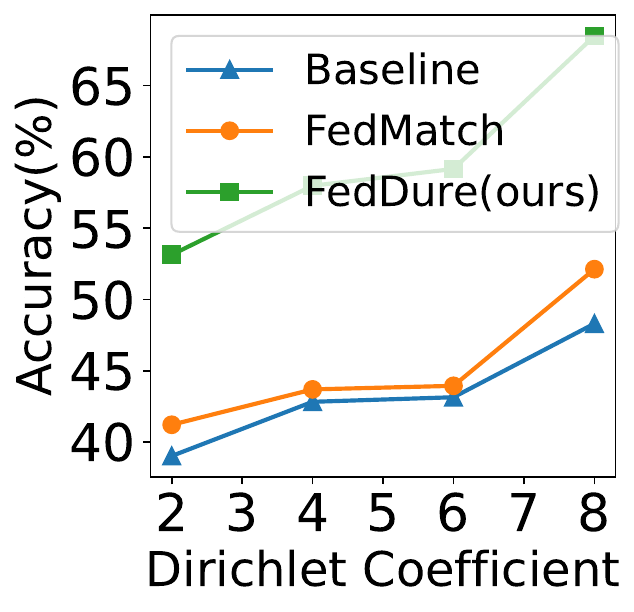}
		\end{minipage}%
	}
	\hfill%
	\centering
	\caption{Impact of different Dirichlet coefficients under (IID, DIR) and (DIR, DIR) settings on CIFAR10 dataset.}
	\label{dir_coefficient}
\end{figure}

\subsection{Performance Comparison}
\label{comparison}

Table \ref{comparasion} reports the overall results of FedDure and other
state-of-the-art methods on the three datasets. These results are averaged over 3 independent runs.
Our FedDure achieves state-of-the-art FSSL performances on all datasets and data settings. 
\textbf{\textit{(IID, IID) setting:}}
compared with the naive combination of FSSL methods and FedMatch, our FedDure significantly outperforms them on all three datasets.
Specifically, when evaluated on CINIC-10, which is a more difficult dataset with a larger amount of unlabeled samples, other methods suffer from the performance bottleneck and are inferior on CIFAR-10 with fewer unlabeled samples. 
These results show that FedDure effectively alleviates the negative influence of mass unlabeled data by regulating the local model's optimization on unlabeled data through knowledge feedback from labeled data using F-reg and C-reg. 
\textbf{\textit{(IID, DIR) setting:}}
our FedDure is slightly affected by weak class mismatch on unlabeled data, but it significantly outperforms by FedMatch 15.26\% on CIFAR10 dataset.
Also, competitive performance is achieved compared to the supervised method FedAvg-SL on Fashion-MNIST.
\textbf{\textit{(DIR, DIR) setting:}}
Under this more challenging and realistic setting,
our FedDure significantly outperforms others by at least 11\% on CIFAR-10 and CINIC-10 datasets.
In particular, the performance of other methods drops dramatically, and in CIFAR10 and Fashion-MNIST datasets, some semi-supervised algorithms are even worse than FedAvg*.
It means that unlabeled data might hurt performance
due to the distribution mismatch between labeled and unlabeled data.
\begin{figure*}[t]
  \centering
  \subfigure[Setting: (IID, DIR) \label{label_ratio_iiddir}]{\includegraphics[width=0.25\linewidth]{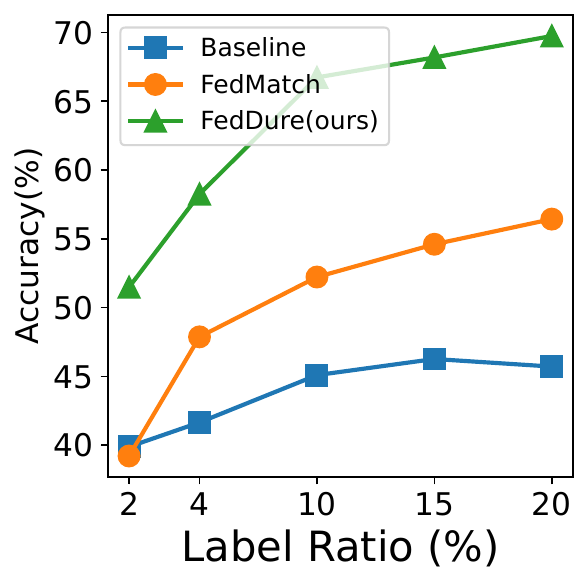}}
  \subfigure[Setting: (DIR, DIR) \label{label_ratio_dirdir}]{\includegraphics[width=0.25\linewidth]{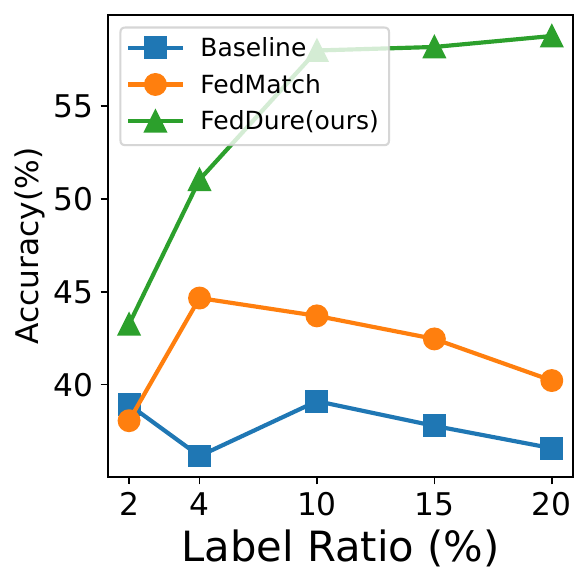}}
  \subfigure[CIFAR-10 \label{clients_cifar}]
  {\includegraphics[width=0.24\linewidth]{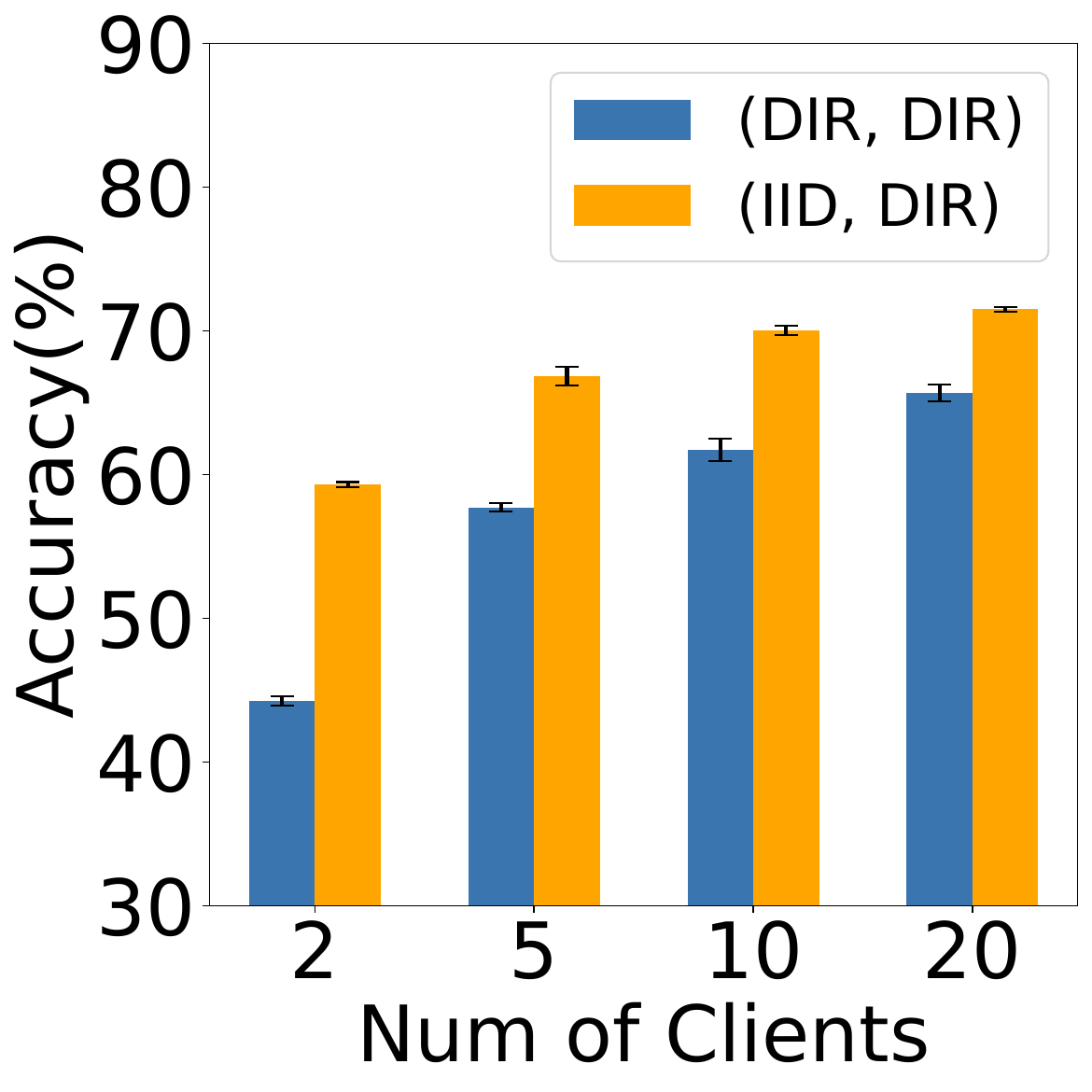}}
  \subfigure[Fashion-MNIST \label{clients_fash}]{\includegraphics[width=0.24\linewidth]{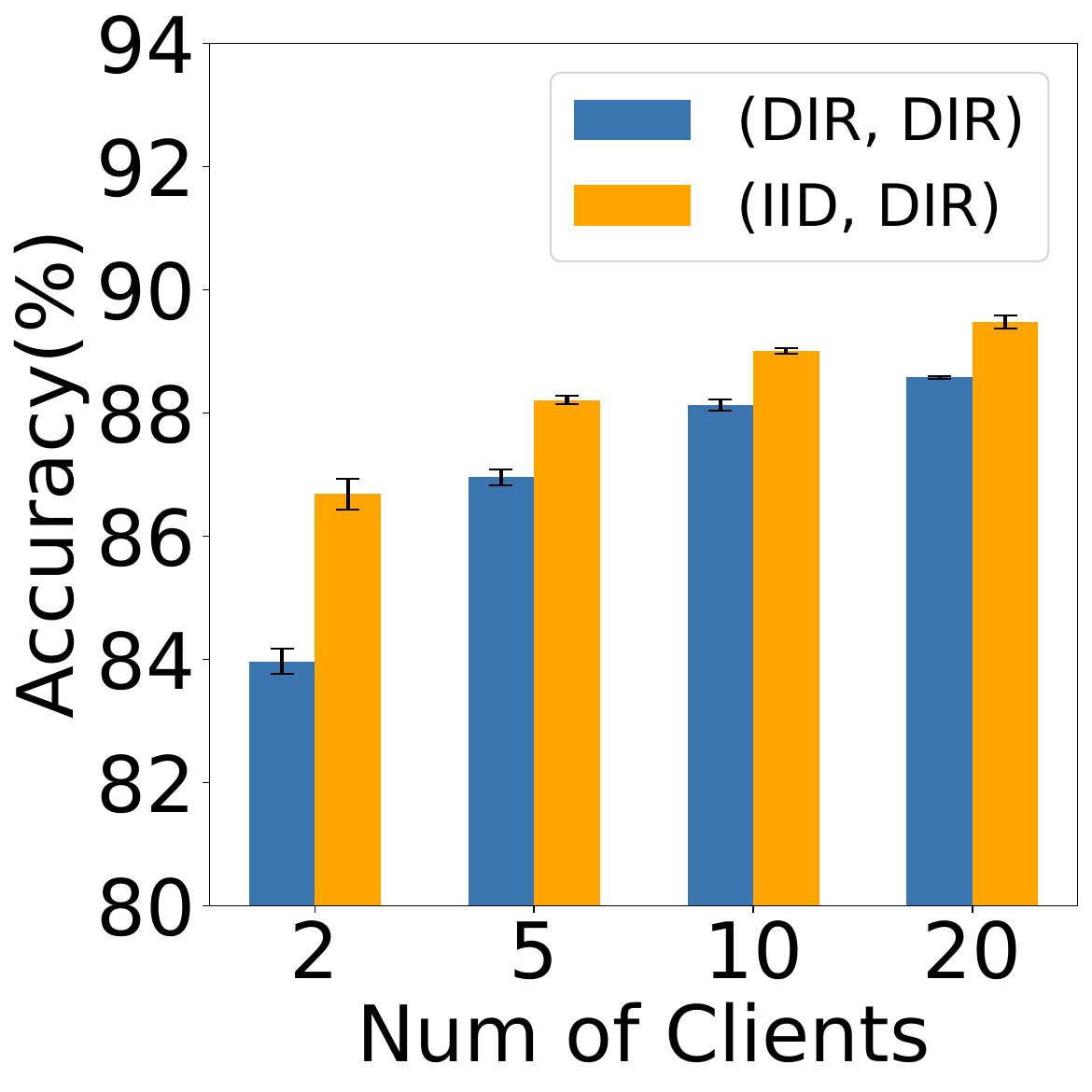}}
 \caption{Analysis of the impacts of the number of labeled data and selected clients. (a) and (b) illustrate that FedDure consistently outperforms FedMatch and Baseline (FedAvg-Fixmatch) using different percentages of labeled data. (c) and (d) show that FedDure scales with increasing numbers of selected clients on CIFAR-10 and Fashion-MNIST datasets.}
 \label{fig_ratio_num}
\end{figure*}

\begin{table}[t]
\centering
			
   \resizebox{1.0\hsize}{!}{
   \begin{tabular}{c|c|c|c}
				\toprule[1.3pt]
				\multirow{2}{*}{Ablated components}    & \multicolumn{3}{c}{CIFAR-10} \\
				\cline{2-4} 
				    & {(DIR, DIR)} & {(IID, DIR)} & {(IID, IID)} \\
				\hline 
				Baseline     &{39.95} &{46.67} &{47.60}   \\
                    Baseline+MAML     &{47.69} &{56.05} &{59.78}   \\  
                 F-reg    &{54.79}  &{64.98} &{65.41 } \\
                 Avg F-reg  &{50.08} &{58.26} &{59.13} \\
                  
                C-reg    &{56.46} &{65.92} &{66.54}   \\
                Avg C-reg  &{52.32}  &{60.07} &{62.07 }    \\
				FedDure    &\textbf{{57.73}} &\textbf{{66.85}} &\textbf{{67.69}} \\			
				\bottomrule[1.3pt]
			\end{tabular}
   }
       \tabcaption{Quantitative analysis of components of FedDure on CIFAR-10 and Fashion-MNIST datasets.}
             \label{tab_components}
\end{table}

\subsection{Ablation Study}

\textbf{Effectiveness of Components.}
To measure the importance of proposed components in our FedDure, we conduct ablation studies with the following variants in Table \ref{tab_components}.
(1) Baseline: the naive combination of FedAvg \cite{mcmahan2017communication} and Fixmatch \cite{sohn2020fixmatch}. 
(2) Baseline+MAML:  an adaptive optimization for baseline based on vanilla meta-learning. The performance improvement over the baseline verifies the insights and the effectiveness of adopting client-specific optimization strategies via meta-learning.
(3) Ours \textit{w} F-reg: this variant denotes our FedDure removes the C-reg (i.e. $g_d$ in Eqn.\ref{eq_total_gradient}) and updates F-reg with the local model.
(4) Ours \textit{w} C-reg: this variant indicates our FedDure replaces the dynamic weight (i.e. $g_u$ in Eqn.\ref{eq_total_gradient}) and uses a fixed threshold to filter low-confidence pseudo labels.
The performance advantage over Baseline+MAML shows that the two components are both more effective for FSSL scenarios. Moreover, C-reg can further make a performance boost under almost all data sets on CIFAR-10. This is because C-reg has
targeted overall knowledge in local training, but there is no significant difference between the two components.
(5) Avg C-reg and Avg F-reg: we set F-reg and C-reg to the average of corresponding regulators in previous rounds. Compared with F-reg and C-reg, the decrease in performance suggests the importance of online client-specific adaptive optimization in dual regulators. 



\textbf{Impacts of Data Heterogeneity.} \;
To demonstrate the robustness of our method against data imbalance, we characterize different levels of imbalances by Dirichlet distribution with different coefficients $\left \{0.3, 0.5, 0.7, 1.0\right\}$ and evaluate multiple methods.
As illustrated in Figure \ref{fig_dir_IIDDIR} and \ref{fig_dir_DIRDIR}, our FedDure consistently showcases substantial performance improvements across different levels of data imbalances. However,  FedMatch and baseline (FedAvg-Fixmatch) suffer from rapid performance degradation when confronted with the higher data heterogeneous \textit{(small coefficient)}


\textbf{Number of Label Data per Client.} We evaluate FedDure under the different percentages of labeled instances in each client in $\left\{2\%, 4\%, 10\%, 15\%, 20\% \right\}$. As illustrated in Figure \ref{label_ratio_iiddir} and \ref{label_ratio_dirdir}, FedDure gains steady performance improvements with the number of labeled data increases in two data settings. In contrast, the baseline's performance remains relatively stagnant across both scenarios. As for FedMatch, a noticeable decline becomes obvious when the labeling ratio exceeds 4\% in the (DIR, DIR) setting. These insightful findings underscore the efficacy of our dual regulators.


\textbf{Number of Selected Clients per Round.} Lastly, we investigate the performance on the impact of the number of selected clients varied in $\left\{ 2,5,10,20\right\}  $. As shown in Figure \ref{clients_cifar} and \ref{clients_fash}, significant improvements can be achieved by increasing the selected clients. Nevertheless, performance gains plateau as the chosen clients surpass a certain threshold. Our contention is that while the number of selected clients does exhibit a positive correlation with overall performance, our method delves into the intrinsic knowledge of each client to enhance the central server's overall performance. In scenarios where there are ample clients, our approach assimilates comprehensive knowledge, resulting in saturated performance.

\section{Conclusion}
In this paper, we introduce a more practical and challenging scenario of FSSL, data distribution is different across clients (external imbalance) and within a client (internal imbalance). We then design a new federated semi-supervised learning framework with dual regulators, FedDure, to address the challenge. Particularly, we propose a coarse-grained regulator (C-reg) to regularize the gradient update in client model training and present a fine-grained regulator (F-reg) to learn an adaptive weighting scheme for unlabeled instances for gradient update. Furthermore, we formulate the learning process in each client as bi-level optimization that optimizes the local model in the client adaptively and dynamically with these two regulators. Theoretically, we show the convergence guarantee of the regulators. Empirically, extensive experiments demonstrate the significance and effectiveness of FedDure. 

\section{Acknowledgments}
This research was supported by fundings from the Key-Area Research and Development Program of Guangdong Province (No. 2021B0101400003), Hong Kong RGC Research Impact Fund (No. R5060-19, No. R5034-18), Areas of Excellence Scheme (AoE/E-601/22-R), General Research Fund (No. 152203/20E, 152244/21E, 152169/22E, 152228/23E), Shenzhen Science and Technology Innovation Commission (JCYJ20200109142008673).

\bibliography{aaai24}

\end{document}